# Distribution Preserving Multiple Hypotheses Prediction for Uncertainty Modeling

Tobias Leemann[1], Moritz Sackmann[1], Jörn Thielecke[1] and Ulrich Hofmann[2]

1- University of Erlangen-Nürnberg
Institute of Information Technology (Communication Electronics)
91058 Erlangen - Germany

2- AUDI AG - Predevelopment of Automated Driving
85045 Ingolstadt - Germany

**Abstract**. Many supervised machine learning tasks, such as future state prediction in dynamical systems, require precise modeling of a forecast's uncertainty. The Multiple Hypotheses Prediction (MHP) approach addresses this problem by providing several hypotheses that represent possible outcomes. Unfortunately, with the common $l_2$ loss function, these hypotheses do not preserve the data distribution's characteristics. We propose an alternative loss for distribution preserving MHP and review relevant theorems supporting our claims. Furthermore, we empirically show that our approach yields more representative hypotheses on a synthetic and a real-world motion prediction data set. The outputs of the proposed method can directly be used in sampling-based Monte-Carlo methods.

## 1 Introduction

Machine learning is a highly popular method for various prediction tasks such as vehicle trajectories [1], video frame prediction [2] or energy demand forecasting [3]. An accurate model of the inherent, aleatoric uncertainty of the prediction is often crucial.

A common approach to characterize the uncertainty is to predict a continuous distribution, e.g., in Mixture Density Networks (MDN) [4]. However, in some cases, it is beneficial to represent the uncertainty through a discrete set of hypotheses. For example, when learning a dynamics model of a complex and non-linear system, exact probabilistic inference is intractable and the sample approximation is used instead. Possible future system states are represented by a set of random particles from the distribution and can be used for Monte-Carlo simulation, e.g., in [1]. A network that directly predicts multiple hypotheses eliminates the need for a stochastic sampling step. Moreover, a well-arranged set of hypotheses can be a better representation of an underlying density than an equally sized set of random samples from that distribution.

In this paper, we name the desired properties of such hypotheses sets and provide precise definitions. Our main contribution is a new loss function that enables a neural network to approximately meet these requirements and output multiple hypotheses, more accurately representing the underlying data density.

## 2 Multiple Hypotheses Prediction (MHP)

Classical machine learning models map a feature vector $\boldsymbol{x} \in \mathbb{R}^m$ to a label vector $\boldsymbol{y} \in \mathbb{R}^n$ using a parametric function $\boldsymbol{f}$, e.g., a neural network (NN), such that $\boldsymbol{f}(\boldsymbol{x}) = \boldsymbol{y}$. The Multiple Hypotheses Prediction (MHP) approach proposed by Rupprecht et al. [5] features a model, which yields a multitude of $N$ possible outcomes $\boldsymbol{y}_i, i = 1, \ldots, N$ for a single input feature vector $\boldsymbol{x}$ instead,

$$\boldsymbol{f}(\boldsymbol{x}) = [\boldsymbol{y}_1, \boldsymbol{y}_2, \ldots, \boldsymbol{y}_N].$$

Training such a model requires a loss function to be minimized. As usual, the training data consists of $K$ tuples of the form $\left(\boldsymbol{x}^{(k)}, \boldsymbol{y}^{(k)}\right)_{k=1,\ldots,K}$. Therefore, a loss function is needed that assigns a cost value to a vector of output hypotheses $\boldsymbol{f}\left(\boldsymbol{x}^{(k)}\right)$ of the model while relying only on the single observation $\boldsymbol{y}^{(k)}$. The authors of [5, 6] propose using the distance to the closest hypothesis as loss

$$\mathcal{L}\left(\boldsymbol{f}(\boldsymbol{x}^{(k)}),\ \boldsymbol{y}^{(k)}\right) = \min_{i=1\ldots N} d\left(\boldsymbol{y}_i(\boldsymbol{x}^{(k)}),\ \boldsymbol{y}^{(k)}\right),$$

where $d$ is an arbitrary distance function. This loss function is sometimes referred to as Winner-Takes-it-All (WTA) loss [7], because only the best hypothesis defines the loss. It implicitly prevents hypotheses from collapsing into a single point and encourages a diverse set of hypotheses. With a specific function $d$, we denote it as $d$-WTA loss. Minimizing the loss on the entire training set is equivalent to optimizing the expected distance to the closest hypothesis. A common choice for $d$ is the squared error, also known as $l_2$-loss, which is proposed in the original paper [5] and used in applications of the method [7].

## 3 The Distribution Preserving Approach

Although the WTA loss can be used with any distance function, its choice will have a defining impact on the arrangement of the resulting hypotheses. For use in Monte-Carlo algorithms, we regard the following two properties as desirable: First, the hypotheses must preserve characteristics of the underlying label distribution $P(\boldsymbol{y})$, e.g., expectation and higher moments, at least in the limiting case $N \to \infty$. We omit the condition on the input $\boldsymbol{x}$ at first. A more formal description is provided in Property 3.1, which is adapted from [8, Theorem 7.5].

**Property 3.1** *Let a stochastic label $\boldsymbol{y}$ have a continuous probability measure $P$ and let $(Y_N)_{N \in \mathbb{N}}, Y_N \subset \mathbb{R}^n, |Y_N| = N$ be a series of sets with each set representing $N$ discrete hypotheses returned by a discretization method of $P$ on $\mathbb{R}^n$. The method is* ***distribution preserving*** *if and only if for every continuous, bounded test function $b : \mathbb{R}^n \to \mathbb{R}$ the following limit is satisfied:*

$$\lim_{N \to \infty} \frac{1}{N} \sum_{\boldsymbol{y}_i \in Y_N} b(\boldsymbol{y}_i) = \mathbb{E}_{\boldsymbol{y} \sim P}\left[b(\boldsymbol{y})\right].$$

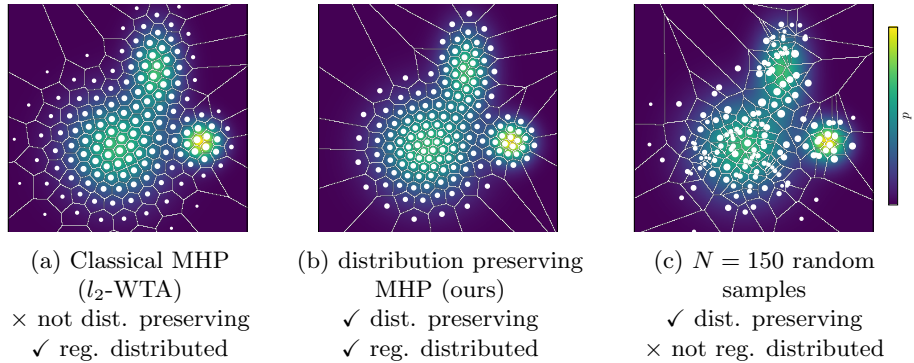

(a) Classical MHP
($l_2$-WTA)
× not dist. preserving
✓ reg. distributed

(b) distribution preserving
MHP (ours)
✓ dist. preserving
✓ reg. distributed

(c) $N = 150$ random
samples
✓ dist. preserving
× not reg. distributed

Fig. 1: Sets of $N = 150$ hypotheses (white dots) for the density $p$ of $P$. The size of the dots is proportional to the amount of data in the corresponding Voronoi regions. With $l_2$-WTA (a), the hypotheses are spread out too far, i.e. the hypotheses' variance does not match that of the distribution. Our method (b) aims to be both *distribution preserving* and *regularly distributed*, whereas random samples (c) are not regularly distributed.

Note that this definition is equal to the weak convergence of measures by the Portmanteau theorem, but we chose this formulation as it is closer to our understanding of distributional properties. Second, the hypotheses need to be *regularly distributed*, i.e., equally spaced locally and forming a regular pattern.

**Property 3.2** *Let $(Y_N)_{N \in \mathbb{N}}$ be a series of sets with $N$ discrete hypotheses by a discretization method. The method yields **regularly distributed** hypotheses if and only if for $N \to \infty$, the shape and volume of Voronoi regions defined by neighboring hypotheses tend to vary only infinitesimally.*

In practice, this property is hard to verify and can only be proven for very special cases (e.g., [9] claims that Voronoi cells form regular hexagons with $l_2$-WTA in two dimensions). Nevertheless, if the hypotheses fulfill properties 3.1 and 3.2, for continuous densities there is an important implication: For a sufficiently large number $N$, each Voronoi cell $V_{i,N}, i = 1, ..., N$ will approximately contain an equal share of data, i.e. $\int_{V_{i,N}} dP \approx \frac{1}{N}$. This can be seen as follows: Consider a small ball $\mathcal{B} \subset \mathbb{R}^n$. The distribution preserving property enforces that the hypotheses are correctly distributed globally, so $B$ will approximately contain $J \approx P(\mathcal{B})N$ hypotheses (use indicator $b = I_{\mathcal{B}}$ in 3.1. to see this). If $\mathcal{B}$ is sufficiently small, the density $p$ will be almost constant on $\mathcal{B}$ due to its continuity. The regular distribution property ensures that the space is equally partitioned among the $J$ hypotheses in the ball locally for a sufficiently large $J$, resulting in a probability share of approximately $\frac{P(\mathcal{B})}{J} \approx \frac{1}{N}$ for each hypothesis.

Figure 1a shows the optimal placement of hypotheses on a density with the aforementioned $l_2$-WTA loss function. This was achieved by directly minimizing the expected loss value on samples of the depicted distribution without use of a NN. It becomes evident that the desired properties cannot be fulfilled because the shares of data for each hypothesis differ by magnitudes. While the definitions and the implication summarize our thoughts, the fact that the $l_2$-WTA loss is not distribution-preserving is already established. If $p$ is the data density, the resulting hypotheses in $n$ dimensions will distribute according to a distribution proportional to $p^{\frac{n}{2+n}}$ and not to $p$ itself [8, Theorem 7.5]. To overcome this, we propose training the MHP model with a different loss function. Specifically, the logarithmic distance

$$l_{dp}(\boldsymbol{a}, \boldsymbol{b}) = \log\left(\|\boldsymbol{a} - \boldsymbol{b}\| + \delta\right), \delta \ll 1$$

("dp" for distribution preserving) preserves non-conditional distributions for $N \to \infty$ when used with the WTA loss as claimed in [10] for continuous probability distributions under mild regularity requirements. $\delta$ is a small constant to overcome numerical issues. Figure 1b shows the distribution of hypotheses that is obtained if the $l_{dp}$-WTA is optimized. Intuitively, the logarithmic distance does not penalize outliers much compared to the $l_2$ distance, which draws the hypotheses towards outer but improbable points. This results in a hypotheses set that is overly spread out for the $l_2$-WTA. Our proposed loss is not so sensitive to outliers and avoids this problem.

The optical impression and the evenly distributed shares of the data indicate that our loss function also yields hypotheses which are highly regularly distributed. Nonetheless, for the scope of this work, we do not provide a general proof. Our contribution is to extend the result of [10] by using the proposed loss function for backpropagation through a neural network and training of an MHP model that can then represent hypotheses conditioned on a feature vector $\boldsymbol{x}$.

## 4 Experiments

To validate the theoretical results and show the benefits of our method, we train a distribution preserving MHP model on an artificial data set and on real-world data and compare it to hypotheses that were obtained with the $l_2$-WTA loss.

First, as an artificial example, we consider a call center and suppose an employee asks every fifth costumer about his satisfaction with the service. Therefore, we model the waiting time $y \in \mathbb{R}_+$ that passes between five successive calls dependent on the time of day $x$. For simplicity, we assume it to be Erlang distributed with the conditional probability

$$p(y|x) = \text{Erlang}\left(\lambda(x), 5\right).$$

The interarrival times between successive calls are exponentially distributed with rate $\lambda(x)$ sinusoidally varying over the time of day $x$ (see Figure 2a). We train an MHP model with samples from the distribution, where first a time $x$ is sampled uniformly on the interval $[6, 20]$ and $y$ is subsequently drawn from the conditional distribution. Our code is available at `https://github.com/tleemann/dpmhp`. By using the $l_{dp}$-WTA loss and backpropagating it through the network, we obtain the hypotheses $y_i(x)$ shown in Figure 2b. They are spaced according to the

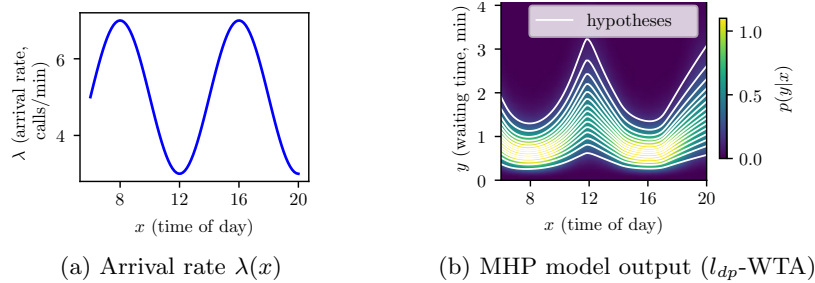

(a) Arrival rate $\lambda(x)$        (b) MHP model output ($l_{dp}$-WTA)

Fig. 2: Arrival rate and hypotheses obtained for the call center example.

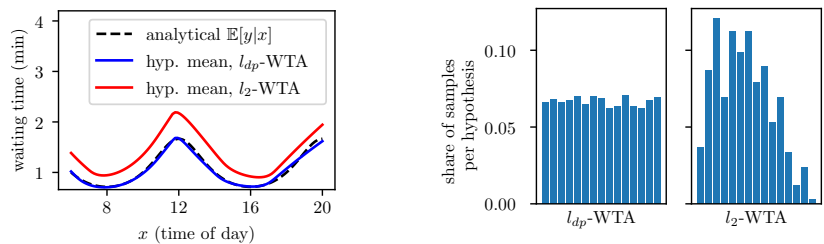

(a) Mean values of the hypotheses and analytical mean of the distribution

(b) Shares of data closest to each of the 15 hypotheses (one bar per hypothesis)

Fig. 3: Further statistics computed on the models for the call center example.

data distribution, with high-density areas being covered denser. Figure 3 provides additional measurements. The distribution parameters as the conditional expectation $\mathbb{E}[y|x]$ are well-preserved, whereas that value differs considerably when using another model with identical hyperparameters trained with the $l_2$-WTA loss (Figure 3a). We observed a similar result for $\text{Var}[y|x]$ (shown in our code example). When drawing a large number of samples ($10^5$) and assigning them to the closest hypothesis, the shares are approximately equal for the proposed method and vary highly when using $l_2$-WTA, indicating that the properties defined are not fulfilled (Figure 3b). Additional measurements and different hypothesis counts can be found in our code.

To showcase the applicability of our approach to real-world data sets, we learn a behavior model for traffic participants, enabling the prediction of future trajectories inspired by [1]. The task features an 18-dimensional feature vector $\boldsymbol{x} \in \mathbb{R}^{18}$ describing the current traffic situation from a specific car's perspective and contains, e.g., its velocity and the distance to a preceding vehicle. Out of these features, the goal is to predict labels $\boldsymbol{y} \in \mathbb{R}^2$ with the vehicle's next action, a steering angle and a linear acceleration. We learn from approximately $K = 3 \cdot 10^6$ tuples $(\boldsymbol{x}, \boldsymbol{y})$ with 100 hypotheses.

We calculate the negative log-likelihood (NLL) of observations in the test set with respect to a density estimated by the set of hypotheses. We use two den-sity estimators, a single normal distribution (Norm) and a kernel density estimator (KDE). The results in Table 1 show that the distribution of the test set is represented more accurately with the proposed $l_{dp}$-WTA.

## 5   Conclusion and Outlook

We introduce a novel approach to generate a set of diverse hypotheses to represent uncertainty in supervised learning problems. In contrast to the literature, our method aims to preserve the characteristics of the underlying data distribution. As a consequence, the resulting hypotheses provide insights on the conditional distribution and can be directly used in sampling-based algorithms such as particle filtering. Further research can determine the impact on long-term prediction of dynamical systems and in other surroundings such as Markov Decision Processes.

| Model, N=100 | NLL (Norm) | NLL (KDE) |
|---|---|---|
| $l_2$-WTA | -0.80 | -0.73 |
| $l_{dp}$-WTA | **-1.44** | **-1.33** |

Table 1: Results on the motion prediction data set. NLL is normalized by the number of samples in the test set. Lower values indicate better fit.

**Acknowledgment:** This work is a result of the research project @CITY – Automated Cars and Intelligent Traffic in the City. The project is supported by the Federal Ministry for Economic Affairs and Energy (BMWi), based on a decision taken by the German Bundestag. The author is solely responsible for the content of this publication.

## References

[1] J. Schulz, C. Hubmann, N. Morin, J. Löchner, and D. Burschka. Learning interaction-aware probabilistic driver behavior models from urban scenarios. In *IEEE Intelligent Vehicles Symposium (IV)*, pages 1326–1333. IEEE, 2019.

[2] C. Finn, I. Goodfellow, and S. Levine. Unsupervised learning for physical interaction through video prediction. In *Advances in Neural Information Processing Systems (NIPS)*, pages 64–72, 2016.

[3] C. Deb, F. Zhang, J. Yang, S. E. Lee, and K. W. Shah. A review on time series forecasting techniques for building energy consumption. *Renewable and Sustainable Energy Reviews*, 74:902–924, 2017.

[4] C. M. Bishop. Mixture density networks. 1994.

[5] C. Rupprecht, I. Laina, R. DiPietro, M. Baust, F. Tombari, et al. Learning in an uncertain world: Representing ambiguity through multiple hypotheses. In *The IEEE International Conference on Computer Vision (ICCV)*, pages 3591–3600, 2017.

[6] A. Guzman-Rivera, D. Batra, and P. Kohli. Multiple choice learning: Learning to produce multiple structured outputs. In *Advances in Neural Information Processing Systems (NIPS)*, pages 1799–1807, 2012.

[7] O. Makansi, E. Ilg, O. Cicek, and T. Brox. Overcoming limitations of mixture density networks: A sampling and fitting framework for multimodal future prediction. In *The IEEE Conference on Computer Vision and Pattern Recognition (CVPR)*, pages 7144–7153, 2019.

[8] S. Graf and H. Luschgy. *Foundations of quantization for probability distributions.* Springer, 2000.

[9] Q. Du, V. Faber, and M. Gunzburger. Centroidal voronoi tessellations: Applications and algorithms. *SIAM review*, 41(4):637–676, 1999.

[10] A. Krishna, S. Mak, and R. Joseph. Distributional clustering: A distribution-preserving clustering method. *arXiv preprint arXiv:1911.05940*, 2019.